\documentclass[runningheads]{llncs}
\usepackage{graphicx}
\usepackage{subcaption}
\usepackage{fontawesome}
\usepackage{booktabs}

\usepackage{color}
\usepackage{todonotes}

\newcommand{\corpuscw}{CT-CWC-18\,}

\begin{document}
\title{Overview of the CLEF-2018 CheckThat! Lab\\ on Automatic Identification and Verification\\ of Political Claims. Task 1: Check-Worthiness\thanks{This paper focuses on Task 1 (Check-Worthiness). For Task 2 (Factuality), see~\cite{clef2018checkthat:task2}.}}

\titlerunning{Overview of the CLEF-2018 CheckThat! Lab. Task 1: Check-Worthiness}
%
\author{%
Pepa Atanasova\inst{1}
\and 
Llu\'{i}s M\`{a}rquez\inst{2} 
\and
Alberto Barr\'{o}n-Cede\~no\inst{3}	
\and	 \\
Tamer Elsayed\inst{4}
\and 
Reem Suwaileh\inst{4}
\and 
Wajdi Zaghouani\inst{5}
\and  \\
Spas Kyuchukov\inst{6} 
\and 
Giovanni Da San Martino\inst{3}
\and 
Preslav Nakov\inst{3}
}
\authorrunning{P. Atanasova et al.}
%
\institute{%
SiteGround, Sofia, Bulgaria \\
\email{pepa.gencheva@siteground.com}\\
\and
Amazon, Barcelona, Spain \\
\email{lluismv@amazon.com}
\and
Qatar Computing Research Institute, HBKU, Doha, Qatar \\
\email{\{albarron, gmartino, pnakov\}@qf.org.qa}\\
\and
Computer Science and Engineering Department, Qatar University, Doha, Qatar \\
\email{\{telsayed, reem.suwaileh\}@qu.edu.qa}	\and
College of Humanities and Social Sciences, HBKU, Doha, Qatar \\
\email{wzaghouani@hbku.edu.qa}\and
Sofia University ``St Kliment Ohridski'', Sofia, Bulgaria \\
\email{spas.kyuchukov@gmail.com}\\
}
\maketitle              
\begin{abstract}
We present an overview of the CLEF-2018 CheckThat! Lab on Automatic Identification and Verification of Political Claims, with focus on Task 1: Check-Worthiness. The task asks to predict which claims in a political debate should be prioritized for fact-checking. In particular, given a debate or a political speech, the goal was to produce a ranked list of its sentences based on their worthiness for fact checking. We offered the task in both English and Arabic,  based on debates from the 2016 US Presidential Campaign, as well as on some speeches during and after the campaign. A total of 30 teams registered to participate in the Lab and seven teams actually submitted systems for Task~1. The most successful approaches used by the participants relied on recurrent and multi-layer neural networks, as well as on combinations of distributional representations, on matchings claims' vocabulary against lexicons, and on measures of syntactic dependency. The best systems achieved mean average precision of 0.18 and 0.15 on the English and on the Arabic test datasets, respectively. This leaves large room for further improvement, and thus we release all datasets and the scoring scripts, which should enable further research in check-worthiness estimation.

\end{abstract}

\section{Introduction}

The current coverage of the political landscape in both the press and in social media has led to an unprecedented situation. Like never before, a statement in an interview, a press release, a blog note, or a tweet can spread almost instantaneously across the globe. This proliferation speed has left little time for double-checking claims against the facts, which has proven critical in politics, e.g.,~during the 2016 US Presidential Campaign, which was influenced by fake news in social media and by false claims. Indeed, some politicians were fast to notice that when it comes to shaping public opinion, facts were secondary, and that appealing to emotions and beliefs worked better, especially in social media. It has been even proposed that this was marking the dawn of a post-truth age.

As the problem became evident, a number of fact-checking initiatives have started, led by organizations such as FactCheck
and Snopes,
among many others. Yet, this has proved to be a very demanding manual effort, which means that only a relatively small number of claims could be fact-checked.\footnote{Full automation is not yet a viable alternative, partly because of limitations of the existing technology, and partly due to low trust in such methods by human users.} This makes it important to prioritize the claims that fact-checkers should consider first. Task 1 of the CheckThat! Lab at CLEF-2018~\cite{clef2018checkthat:overall} aims to help in that respect,
asking participants to build systems that can mimic the selection strategies of a particular fact-checking organization: \url{factcheck.org}.
It is defined as follows: 

\begin{center}
\parbox[c]{.7\textwidth}{
\textit{Given a transcription of a political debate/speech, predict which claims should be prioritized for fact-checking. 
}}
\end{center}

\begin{figure}[h!]
\centering
\includegraphics[width=.9\textwidth]{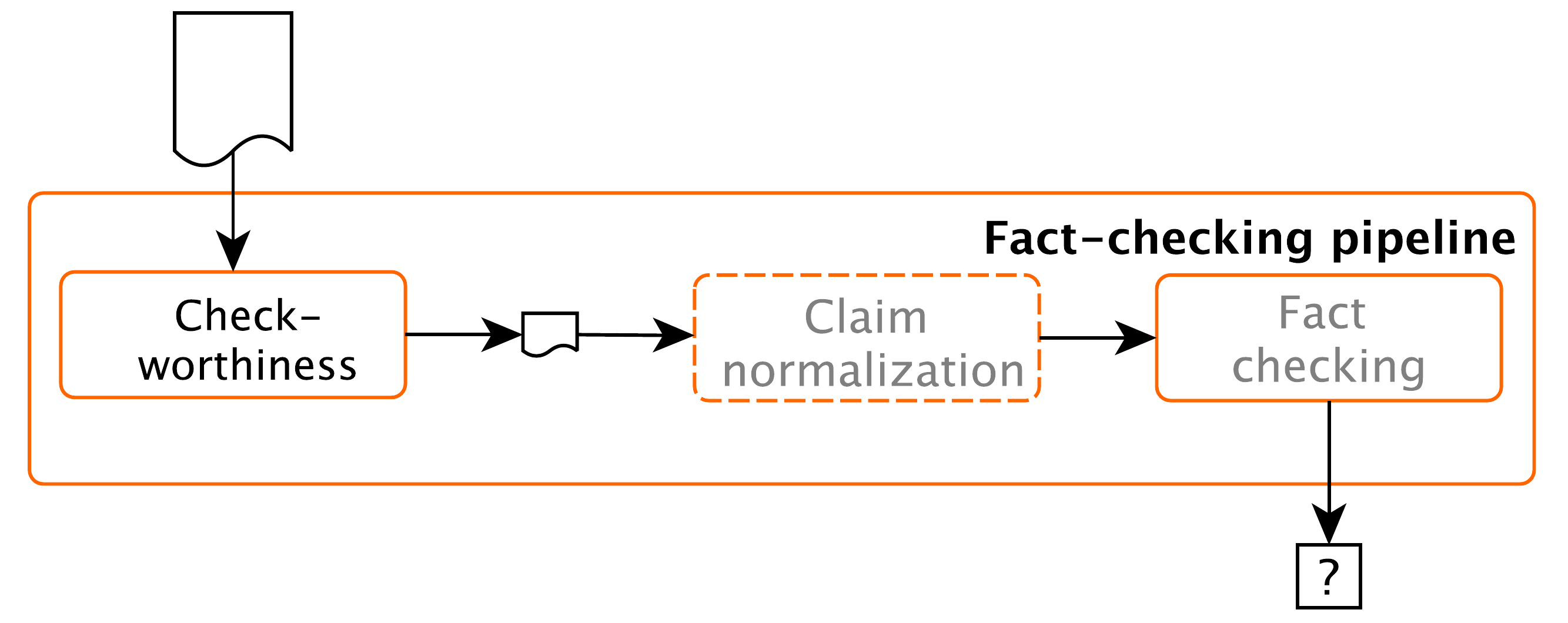}
 \caption{\label{fig:pipeline}The general fact-checking pipeline. First, the input document is analyzed to identify sentences containing check-worthy claims (this task), then these claims are extracted and normalized, and finally they are fact-checked.}
\end{figure}

The goal is to produce a \textit{ranked list} of sentences ordered by their worthiness for fact-checking. This is the first step in the pipeline of the full fact-checking process, displayed in Figure~\ref{fig:pipeline}. Refer to~\cite{clef2018checkthat:task2} for details on the fact-checking task.

\begin{figure}[tbh]
\centering

\begin{subfigure}{\textwidth}
\centering
\begin{tabular}{p{28mm} p{80mm} @{\hspace{5mm}}p{5mm}}
\toprule
Hillary Clinton:	& I think my husband did a pretty good job in the 1990s.	& 	\\
Hillary Clinton:	& I think a lot about what worked and how we can make it work again\ldots	& 	\\
Donald Trump:	& Well, he approved NAFTA...	& \faCheckCircleO	\\
\bottomrule
\end{tabular}         
\caption{Fragment from the First 2016 US Presidential Debate.}
\label{fig:examples-a}
\end{subfigure}

\begin{subfigure}{\textwidth}
\centering
\begin{tabular}{p{28mm} p{80mm} @{\hspace{5mm}} p{5mm}}
\toprule
Hillary Clinton:	& Take clean energy	& \\
Hillary Clinton:	& Some country is going to be the clean-energy superpower of the 21st century.& \\
Hillary Clinton:	& Donald thinks that climate change is a hoax perpetrated by the Chinese.& \faCheckCircleO\\
Hillary Clinton:	& I think it's real.& \\
Donald Trump:	& I did not.& 	\\
\bottomrule
\end{tabular}         
\caption{Another fragment from the First 2016 US Presidential Debate.}
\label{fig:examples-b}
\end{subfigure}
 \caption{\label{fig:examples}English debate fragments: check-worthy sentences are marked with \faCheckCircleO.}
 \end{figure}

\noindent We offered the task in two languages, English and Arabic. 
Figure~\ref{fig:examples} shows examples of English debate fragments. 
In example~\ref{fig:examples-a}, Hillary Clinton discusses the performance of her husband Bill Clinton while he was US president. Donald Trump fires back with a claim that is worth fact-checking: that Bill Clinton approved NAFTA. 
In example~\ref{fig:examples-b}, whether Donald Trump thinks about climate change as charged by Hillary Clinton is also worth fact-checking.

The rest of this paper is organized as follows. Section~\ref{sec:related} discusses related work. Section~\ref{sec:framerork} describes the evaluation framework and the task setup. Section~\ref{sec:approaches} provides an overview of the participating systems, followed by the official results in Section~\ref{sec:results}, and discussion in Section~\ref{sec:discuss}, before we conclude in Section~\ref{sec:conclusions}.

\section{Related Work}
\label{sec:related}

Journalists, online users, and researchers are well aware of the proliferation of false information.
For example, there was a 2016 special issue of the ACM Transactions on Information Systems journal on Trust and Veracity of Information in Social Media \cite{Papadopoulos:2016:OSI}, and
there is a Workshop on Fact Extraction and Verification at EMNLP'2018.
Moreover, there have been several related shared tasks, e.g., a SemEval-2017 shared task on Rumor Detection \cite{derczynski-EtAl:2017:SemEval}, 
an ongoing FEVER challenge on Fact Extraction and VERification at EMNLP'2018, 
the present CLEF'2018 Lab on Automatic Identification and Verification of Claims in Political Debates,
and an upcoming task at SemEval'2019 on Fact-Checking in Community Question Answering Forums.

Automatic fact-checking was envisioned in \cite{vlachos2014fact} as a multi-step process that includes
(\emph{i})~identifying check-worthy statements 
\cite{gencheva-EtAl:2017:RANLP,Hassan:15,NAACL2018:claimrank},
(\emph{ii})~generating questions to be asked about these statements~\cite{karadzhov2017fully},
(\emph{iii})~
retrieving relevant information to create a knowledge base~\cite{shiralkar2017finding}, 
and
(\emph{iv})~inferring the veracity of the statements, e.g., using text analysis~\cite{Castillo:2011:ICT:1963405.1963500,rashkin2017truth}
or external sources~\cite{karadzhov2017fully,Popat:2017:TLE:3041021.3055133}. 

\noindent The first work to target check-worthiness was the ClaimBuster system~\cite{Hassan:15}. It was trained on data that was manually annotated by students, professors, and journalists, where each sentence was annotated as \textit{non-factual}, \textit{unimportant factual}, or \textit{check-worthy factual}. The data consisted of transcripts of historical US election debates covering the period from 1960 until 2012 for a total of 30 debates and 28,029 transcribed sentences. In each sentence, the speaker was marked: candidate vs. moderator.
The ClaimBuster used an SVM classifier and a manifold of features such as sentiment, TF.IDF word representations, part-of-speech (POS) tags, and named entities. It produced a check-worthiness ranking on the basis of the SVM prediction scores.
The ClaimBuster system did not try to mimic the check-worthiness decisions for any specific fact-checking organization; yet, it was later evaluated against CNN and PolitiFact \cite{Hassan2016ComparingAF}. In contrast, our dataset is based on actual annotations by a fact-checking organization, and we release freely all data and associated scripts (while theirs is not available).

More relevant to the setup of Task 1 of this Lab is the work of \cite{gencheva-EtAl:2017:RANLP}, who focused on debates from the US 2016 Presidential Campaign and used pre-existing annotations from nine respected fact-checking organizations (PolitiFact, FactCheck, ABC, CNN, NPR, NYT, Chicago Tribune, The Guardian, and Washington Post): a total of four debates and 5,415 sentences. Beside many of the features borrowed from ClaimBuster ---together with sentiment, tense, and some other features---, their model pays special attention to the context of each sentence. This includes whether it is part of a long intervention by one of the actors and even its position within such an intervention. The authors predicted both (\emph{i})~whether any of the fact-checking organizations would select the target sentence, and also (\emph{ii})~whether a specific one would select it. 

In follow-up work, \cite{NAACL2018:claimrank} developed ClaimRank, which can mimic the claim selection strategies for each and any of the nine fact-checking organizations, as well as for the union of them all. Even though trained on English, it further supports Arabic, which is achieved via cross-language English-Arabic embeddings.

The work of \cite{Patwari:17} also focused on the 2016 US Election campaign, and they also used data from nine fact-checking organizations (but slightly different set from above). They used presidential (3 presidential one vice-presidential) and primary debates (7 Republican and 8 Democratic) for a total of 21,700 sentences.
Their setup asked to predict whether any of the fact-checking sources would select the target sentence. They used a boosting-like model that takes SVMs focusing on different clusters of the dataset and the final outcome was considered as that coming from the most confident classifier. The features considered ranged from LDA topic-modeling to POS tuples and bag-of-words representations.

We follow a setup that is similar to that of \cite{gencheva-EtAl:2017:RANLP,NAACL2018:claimrank,Patwari:17}, but we manually verify the selected sentences, e.g.,~to adjust the boundaries of the check-worthy claim, and also to include all instances of a selected check-worthy claim (as fact-checkers would only comment on one instance of a claim). We further have an Arabic version of the dataset. Finally, we chose to focus on a single fact-checking organization.

\section{Evaluation Framework}
\label{sec:framerork}


\subsection{Data}
\label{sub:datasets}

For Task 1, we produced the \corpuscw dataset,\footnote{\url{http://github.com/clef2018-factchecking/clef2018-factchecking}} which stands for CheckThat! Check-Worthiness 2018 corpus.
It includes transcripts from the 2016 US Presidential campaign, together with some more recent political speeches. In order to derive the annotation, we used the publicly available analysis carried out by FactCheck.org.\footnote{See for example, \url{http://transcripts.factcheck.org/presidential-debate-hofstra-university-hempstead-new-york/}}
We considered those claims whose factuality was challenged by the fact-checkers as check-worthy and we made them positive instances in the dataset.
Note that our annotation is at the sentence level. Therefore, if only part of a sentence was fact-checked, we annotated the entire sentence as a positive example. If a claim spanned more than one sentence, we annotated all these sentences as positive. 
Moreover, in some cases, the same claim was made multiple times in a debate/speech, and thus we annotated all these sentences that referred to it rather than the one that was annotated by the fact-checkers. Finally, we manually refined the annotations by moving them to a neighboring sentence (e.g., in case of argument) or by adding/excluding some annotations.

As shown in Table~\ref{tab:datasets-overview}, the English \corpuscw\  is comprised of five debates and five speeches. To produce Arabic data, we hired translators to translate five debates and Donald Trump's acceptance speech. We released the first three debates as training data, and we used the remaining debates/speeches for testing.

\begin{table}[h]
\centering
\begin{tabular}{c @{\hspace{3mm}} l@{\hspace{3mm}}c@{\hspace{3mm}}c @{\hspace{3mm}}c}
\toprule 
 & \bf Type & \bf Partition & \bf Sent.	& \bf CW \\ 
 \midrule 
 &  \bf Debates				\\
\faLanguage & 1st Presidential	& train	& 1,403	& 37\\
\faLanguage & 2nd Presidential	& train	& 1,303	& 25 \\
\faLanguage & Vice-Presidential	& train	& 1,358 & 28 \\
\hline
\faLanguage & 3rd Presidential	& test		& 1,351 & 77 \\
\faLanguage & 9th Democratic	& test		& 1,464 & 17\\
\midrule
  &  \bf Speeches				\\
\faLanguage & Donald Trump Acceptance& test		& \,\,\,375	& 21 \\
  & Donald Trump at the World Economic Forum & test		& \,\,\,245	& 11 \\
  & Donald Trump at a Tax Reform Event & test	& \,\,\,412	& 16 \\
  &  Donald Trump's Address to Congress & test	& \,\,\,390	& 15 \\
  &  Donald Trump's Miami Speech 	& test	& \,\,\,645	& 35 \\	\bottomrule
\multicolumn{2}{l}{\bf Total English} & & \bf 8,946 & \bf 282\\  
\multicolumn{2}{l}{\bf Total Arabic} & & \bf 7,254 & \bf 205\\  
\bottomrule
\end{tabular}
\caption{\label{tab:datasets-overview} Total number of sentences and those identified as check-worthy (CW) in \corpuscw. The documents available in Arabic are marked with \faLanguage.}
\end{table}

\noindent Note that it was forbidden to use external datasets with fact-checking related annotations. However, it was allowed to extract information from the Web, from Twitter, etc., but the retrieved URLs had to be checked for sanity using a script that we provided to the participants. The script tried to make sure no information from fact-checking websites would be used.

\subsection{Evaluation Measures}

As we shaped this task as an information retrieval problem, in which check-worthy instances should be ranked at the top of the list, we opted for using mean average precision as the official evaluation measure. It is defined as follows:

\begin{equation}
 MAP = \frac{\sum_{d=1}^D AveP(d)}{D}
\end{equation}
where $d\in D$ is one of the debates/speeches, and $AveP$ is the average precision:

\begin{equation}
 AveP = \frac{\sum_{k=1}^K (P(k)\times \delta (k))}{\mbox{\# check-worthy claims}  }
\end{equation}
where $P(k)$ refers to the value of precision at rank $k$ and $\delta(k)=1$ iff the claim at that position is check-worthy. 

Following \cite{gencheva-EtAl:2017:RANLP}, we further report the results for some other measures: (\emph{i})~mean reciprocal rank (MRR), (\emph{ii})~mean R-Precision (MR-P), and (\emph{iii})~mean precision@$k$ (P@$k$). Here \emph{mean} refers to macro-averaging over the testing debates/speeches.

\section{Overview of Participants' Approaches}
\label{sec:approaches}

Table~\ref{tab:approches-summary} offers a summary of the used approaches and representations; see the system description papers for more detail.

\textbf{Prise de Fer}~\cite{T1-Zuo:2018} normalized the texts, e.g., by unifying the speakers' names, and also created additional datasets out of the provided debates by collecting the sentences by a single participant in the debate, thus mimicking speeches. They used averaged word embeddings and bag-of-words representations, after stemming and stopword removal. They also considered the number of negations, verbal forms, as well as clauses and phrases and named entities, among other features. Their prediction model comes in the form of either a multilayer perceptron or a support vector machine. In any case, the decisions made by the model can be overridden by a number of heuristic rules that take into account the length of the intervention or the appearance of certain phrases such as ``thank you'' or a question mark. 

\textbf{Copenhagen}~\cite{T1-Hansen:2018} used a recurrent neural network. Their input consists of a combination of word2vec embeddings~\cite{mikolov-yih-zweig:2013:NAACL-HLT}, part of speech tags, and syntactic dependencies. These representations are fed to a GRU neural network with attention. They further combined their approach with that proposed in~\cite{gencheva-EtAl:2017:RANLP}. This combination boosted their performance on cross-validation, but their neural network alone performed better on the test dataset.

\textbf{bigIR}~\cite{T1-Yasser:2018}used a learning-to-rank approach based on the MART algorithm~\cite{Friedman:01}. Their features are organized in five families: (\emph{i})~word embeddings, and binary features expressing the presence of (\emph{ii})~different types of named entities, (\emph{iii})~part-of-speech tags, (\emph{iv})~sentiment labels, and (\emph{v})~topics. Moreover, they over-sampled the positive instances in the training set in order to alleviate the impact of class imbalance.

\textbf{UPV-INAOE-Autoritas}~\cite{T1-Ghanem:2018} used a $k$-nearest neighbors classifier. Their representation is based on character $n$-grams, after removing irrelevant contents by means of text distortion~\cite{Granados:11}. Regardless of the outcome of the distortion model, words were retained if they were part of named entities or were found in some linguistic lexicons. 


\textbf{RNCC}~\cite{T1-Agez:2018} used support vector machines with different kernels as well as random forests. Their representations are a subset of the values included in the so-called information retrieval nutritional labels of~\cite{Fuhr:18}, which they trained on various datasets.

Two of the participating teams did not submit system description papers, and below we describe their systems based on the limited information that they provided as a short description at system submission time:

The \textbf{fragarach} team, from the Faculty of Mathematics and Informatics, Sofia University, used a linear SVM with a variety of features including averaged word embeddings, sentence length, average length of the words, number of punctuation marks, number of stop words, positive/negative sentiment, and part of speech tags. They further performed feature selection to be able to focus on the most promising words and $n$-grams.

The \textbf{blue} team, from the Indian Institute of Technology Kharagpur, used an LSTM with 100-hidden dimensions with attention, taking the five sentences that preceded the target sentence as context.

\begin{table}[t]
\centering

\begin{tabular}{lc @{\hspace{-0mm}} c @{\hspace{-0mm}} c @{\hspace{-0mm}} c @{\hspace{-0mm}} c}
\toprule
\bf Learning Models	& \cite{T1-Agez:2018}	& \cite{T1-Ghanem:2018}	& \cite{T1-Hansen:2018}	& \cite{T1-Yasser:2018}	& \cite{T1-Zuo:2018}	\\ \midrule
Recurrent neural nets	&			& 			& \faCheckSquare	&			& 	\\			
Multilayer perceptron	&			& 			&			&		 	&\faCheckSquare		\\
Support vector machines	& \faCheckSquare	& 			& 			&			&\faCheckSquare		\\
Random forest		& \faCheckSquare	&			&			&			&		\\
$k$-nearest neighbors	&			& \faCheckSquare	&			&			&		\\
Gradient boosting	&			& 			&			& \faCheckSquare	&		\\	
\midrule	\\

\bf Teams		&	\\
\cite{T1-Agez:2018} RNCC		& & \multicolumn{4}{l}{[--] fragarach}	\\
\cite{T1-Ghanem:2018} UPV-INAOE-Autoritas	& &\multicolumn{4}{l}{[--] blue}\\
\cite{T1-Hansen:2018} Copenhagen	& 	\\
\cite{T1-Yasser:2018} bigIR\\
\cite{T1-Zuo:2018} Prise de Fer\\
\bottomrule
\end{tabular}
\begin{tabular}{l @{\hspace{-1mm}} c @{\hspace{-0mm}} c @{\hspace{-0mm}} c @{\hspace{-0mm}} c @{\hspace{-0mm}} c}
\toprule
\bf Representations	& \cite{T1-Agez:2018}	& \cite{T1-Ghanem:2018}	& \cite{T1-Hansen:2018}	& \cite{T1-Yasser:2018}	& \cite{T1-Zuo:2018}	\\ \midrule
Bag of words		&			& 			&			&		 	& \faCheckSquare	\\
Character $n$-grams		&			& \faCheckSquare	&			&			&		\\
Part of speech tags	&			& 			& \faCheckSquare	& \faCheckSquare 	& \faCheckSquare	\\
Verbal forms		&			& 			&			&			& \faCheckSquare	\\
Negations		&			& 			&			&			& \faCheckSquare\\
Named entities		&			& 			&			& \faCheckSquare 	&  \faCheckSquare	\\	\midrule

Sentiment		&			& 			&			& \faCheckSquare	&  \faCheckSquare		\\
Topics			&			& 			&			& \faCheckSquare	&		\\	\midrule
IR nutritional labels	& \faCheckSquare	&		&		&		&		\\
Clauses			&			& 			&			&			& \faCheckSquare\\
Syntactic dependency &			& 			& \faCheckSquare	&			&  \faCheckSquare \\	\midrule

Word embeddings		&			& 			& \faCheckSquare	& \faCheckSquare	& \faCheckSquare 	\\
\bottomrule

\end{tabular}
\caption{\label{tab:approches-summary}Summary of the models and representations used by the participants.}
\end{table}

\begin{table}[t]
\centering 
\begin{tabular}{l @{\hspace{1mm}} c@{\hspace{1mm}}c@{\hspace{1mm}} c@{\hspace{1mm}}  c@{\hspace{1mm}} c@{\hspace{1mm}}c c@{\hspace{1mm}} c@{\hspace{1mm}} c}
\toprule
		& \bf  MAP	& MRR		& MR-P		& MP@1		& MP@3		& MP@5		& MP@10		& MP@20		& MP@50  \\
\midrule
\multicolumn{2}{l}{\bf Prise de Fer~\cite{T1-Zuo:2018}}	\\
\,\,\,primary	& \bf \underline{.1332$_{(1)}$}	& \bf .4965$_{(1)}$	& \bf .1352$_{(1)}$	& \bf .4286$_{(1)}$	& \bf .2857$_{(1)}$	& .2000$_{(2)}$	& .1429$_{(3)}$	& \bf .1571$_{(1)}$	& .1200$_{(2)}$\\ 
\scriptsize  \,\,\,cont. 1	& \scriptsize .1366	& \scriptsize .5246	& \scriptsize .1475	& \scriptsize .4286	& \scriptsize .2857	& \scriptsize .2286	& \scriptsize .1571	& \scriptsize .1714	& \scriptsize .1229\\ 
\scriptsize \,\,\,cont. 2	& \scriptsize .1317	& \scriptsize .4139	& \scriptsize .1523	& \scriptsize .2857	& \scriptsize .1905	& \scriptsize .1714	& \scriptsize .1571	& \scriptsize .1571	& \scriptsize .1429\\ 
\midrule

\multicolumn{2}{l}{\bf Copenhagen \cite{T1-Hansen:2018}}	\\
\,\,\,primary	& .1152$_{(2)}$	& .3159$_{(5)}$	& .1100$_{(5)}$	& .1429$_{(3)}$	& .1429$_{(4)}$	& .1143$_{(3)}$	& .1286$_{(4)}$	& .1286$_{(2)}$	& \bf .1257$_{(1)}$\\ 
\scriptsize \,\,\,cont. 1	& \scriptsize .1810	& \scriptsize .6224	& \scriptsize .1875	& \scriptsize .5714	& \scriptsize .4286	& \scriptsize .3143	& \scriptsize .2571	& \scriptsize .2357	& \scriptsize .1514\\ 
\midrule

\multicolumn{5}{l}{\bf UPV--INAOE--Autoritas~\cite{T1-Ghanem:2018}} 	\\
\,\,\,primary	& .1130$_{(3)}$	& .4615$_{(2)}$	& .1315$_{(2)}$	& .2857$_{(2)}$	& .2381$_{(2)}$	& \bf .3143$_{(1)}$	& \bf .2286$_{(1)}$	& .1214$_{(3)}$	& .0886$_{(4)}$\\ 
 \scriptsize \,\,\,cont. 1	&  \scriptsize .1232	&  \scriptsize .3451	&  \scriptsize .1022	&  \scriptsize .1429	&  \scriptsize .2857	&  \scriptsize .2286	&  \scriptsize .1429	&  \scriptsize .1143	&  \scriptsize .0771\\ 
 \scriptsize \,\,\,cont. 2	&  \scriptsize .1253	&  \scriptsize .5535	&  \scriptsize .0849	&  \scriptsize .4286	&  \scriptsize .4286	& \scriptsize .2571	& \scriptsize .1429	&  \scriptsize.1286	&  \scriptsize.0771\\ 
\midrule

\multicolumn{2}{l}{\bf bigIR~\cite{T1-Yasser:2018}}	\\
\,\,\,primary	& .1120$_{(4)}$	& .2621$_{(6)}$	& .1165$_{(4)}$	& .0000$_{(4)}$	& .1429$_{(4)}$	& .1143$_{(3)}$	& .1143$_{(5)}$	& .1000$_{(5)}$	& .1114$_{(3)}$\\ 
 \scriptsize \,\,\,cont. 1	&  \scriptsize .1319	&  \scriptsize .2675	&  \scriptsize .1505	&  \scriptsize .1429	&  \scriptsize .0952	&  \scriptsize .0857	&  \scriptsize .1714	&  \scriptsize .1786	&  \scriptsize .1343\\ 
 \scriptsize \,\,\,cont. 2	&  \scriptsize .1116	&  \scriptsize .2195	&  \scriptsize .1294	&  \scriptsize .0000	&  \scriptsize .1429	&  \scriptsize .1429	&  \scriptsize .1857	&  \scriptsize .1429	&  \scriptsize .0886\\ 
\midrule

\multicolumn{2}{l}{\bf fragarach}	\\
\,\,\,primary	& .0812$_{(5)}$	& .4477$_{(3)}$	& .1217$_{(3)}$	& .2857$_{(2)}$	& .1905$_{(3)}$	& .2000$_{(2)}$	& .1571$_{(2)}$	& .1071$_{(4)}$	& .0743$_{(5)}$\\ 
\midrule

\multicolumn{2}{l}{\bf blue}	\\
\,\,\,primary	& .0801$_{(6)}$	& .2459$_{(7)}$	& .0576$_{(7)}$	& .1429$_{(3)}$	& .0952$_{(5)}$	& .0571$_{(4)}$	& .0571$_{(6)}$	& .0857$_{(6)}$	& .0600$_{(6)}$\\ 
\midrule

\multicolumn{2}{l}{\bf RNCC~\cite{T1-Agez:2018}}	\\
\,\,\,primary	& .0632$_{(7)}$	& .3775$_{(4)}$	& .0639$_{(6)}$	& .2857$_{(2)}$	& .1429$_{(4)}$	& .1143$_{(3)}$	& .0571$_{(6)}$	& .0571$_{(7)}$	& .0486$_{(7)}$\\ 
 \scriptsize \,\,\,cont. 1	&  \scriptsize .0886	&  \scriptsize .4844	&  \scriptsize .0945	&  \scriptsize .4286	&  \scriptsize .1429	&  \scriptsize .1714	&  \scriptsize .1286	&  \scriptsize .1000	&  \scriptsize .0714\\ 
 \scriptsize \,\,\,cont. 2	&  \scriptsize .0747	&  \scriptsize .2198	&  \scriptsize .0984	&  \scriptsize .0000	&  \scriptsize .0952	&  \scriptsize .1143	&  \scriptsize .1000	&  \scriptsize .1000	&  \scriptsize .0829\\
 \midrule
\multicolumn{2}{l}{\bf \emph{Baselines}}	\\
\,\,\,n-gram	& .1201 & 	.4087 & 	.1280 & 	.1429 & 	.2857 & 	.1714 & 	.1571 & 	.1357 & 	.1143\\ 
\,\,\,random	& .0485 & 	.0633 & 	.0359 & 	.0000 & 	.0000 & 	.0000 & 	.0286 & 	.0214 & 	.0429\\ 
 \bottomrule
\end{tabular}
\caption{\label{tab:results-checkworthines-en}English results, ranked based on MAP, the official evaluation measure. The best score per evaluation measure is shown in bold.}
\end{table}

\section{Evaluation Results}
\label{sec:results}

The participants were allowed to submit one primary and up to two contrastive runs in order to test variations or alternative models. For ranking purposes, only the primary submissions were considered. A total of seven teams submitted runs for English, and two of them also did so for Arabic. 

\textbf{English.} Table~\ref{tab:results-checkworthines-en} shows the results for English. The best primary submission was that of the \emph{Prise de Fer} team~\cite{T1-Zuo:2018}, which used a multilayer perceptron and a feature-rich representation. We can see that they had the best overall performance not only on the official MAP measure, but also on six out of nine evaluation measures (and they were 2nd or 3rd on the rest).

\noindent Interestingly, the top-performing run for English was an unofficial one, namely the contrastive~1 run by the \emph{Copenhagen} team~\cite{T1-Hansen:2018}. As described in Section~\ref{sec:approaches}, this model consisted of a recurrent neural network on three representations. They submitted a system that combined their neural network with the model of~\cite{gencheva-EtAl:2017:RANLP} as their primary submission, but their neural network alone (submitted as contrastive 1), performed better on the test set. This can be due to the model of \cite{gencheva-EtAl:2017:RANLP} relying on structural information, which was not available for the speeches included in the test set (cf. Section~\ref{sub:datasets}).

To put these results in perspective, the bottom of Table~\ref{tab:results-checkworthines-en} shows the results for two baselines: (\emph{i})~a random permutation of the input sentences, and (\emph{ii})~an n-gram based classifier. We can see that all systems managed to outperform the \emph{random} baseline on all measures by a margin. However, only two runs managed to beat the \emph{n-gram} baseline: the primary run of the \emph{Prise de Fer} team, and the contrastive~1 run of the \emph{Copenhagen} team.

\begin{table}[t]
\centering 
\begin{tabular}{l @{\hspace{1mm}} c@{\hspace{1mm}}c@{\hspace{1mm}} c@{\hspace{1mm}}  c@{\hspace{1mm}} c@{\hspace{1mm}}c c@{\hspace{1mm}} c@{\hspace{1mm}} c}
\multicolumn{2}{c}{}  \\
\toprule
		&   \bf MAP	& MRR	& MR-P	& MP@1	& MP@3	& MP@5	& MP@10		& MP@20	& MP@50  \\
\midrule

\multicolumn{2}{l}{\bf bigIR~\cite{T1-Yasser:2018}}	\\
\,\,\,primary	& \bf .0899$_{(1)}$	& .1180$_{(2)}$	& \bf .1105$_{(1)}$	& .0000$_{(2)}$	& .0000$_{(2)}$	& .0000$_{(2)}$	& \bf .1333$_{(1)}$	& \bf .1000$_{(1)}$	& \bf .1133$_{(1)}$\\ 
\scriptsize \,\,\,cont. 1	& \scriptsize .1497	& \scriptsize .2805	& \scriptsize .1760	& \scriptsize .0000	& \scriptsize .3333	& \scriptsize .3333	& \scriptsize .2667	& \scriptsize .2333	& \scriptsize .1533\\ 
\scriptsize \,\,\,cont. 2	& \scriptsize .0962	& \scriptsize .1660	& \scriptsize .0895	& \scriptsize .0000	& \scriptsize .1111	& \scriptsize .2000	& \scriptsize .1667	& \scriptsize .1000	& \scriptsize .0867\\ 
\midrule

\multicolumn{5}{l}{\bf UPV--INAOE--Autoritas~\cite{T1-Ghanem:2018}}	\\
\,\,\,primary	& .0585$_{(2)}$	& \bf .3488$_{(1)}$	& .0087$_{(2)}$	& \bf .3333$_{(1)}$	& \bf .1111$_{(1)}$	& \bf .0667$_{(1)}$	& .0333$_{(2)}$	& .0167$_{(2)}$	& .0400$_{(2)}$\\ 
\scriptsize \,\,\,cont. 1	& \scriptsize .1168	& \scriptsize .6714	& \scriptsize .0649	& \scriptsize .6667	& \scriptsize .6667	& \scriptsize .4000	& \scriptsize .2000	& \scriptsize .1000	& \scriptsize .0733\\ 
\midrule
\multicolumn{2}{l}{\bf \emph{Baselines}}	\\
\,\,\,n-gram	& .0861 & 	.2817 & 	.0981 & 	.0000 & 	.3333 & 	.2667 & 	.1667 & 	.1667 & 	.0867\\ 
\,\,\,random	& .0460 & 	.0658 & 	.0375 & 	.0000 & 	.0000 & 	.0000 & 	.0333 & 	.0167 & 	.0333\\ 
\bottomrule
\end{tabular}
\caption{\label{tab:results-checkworthines-ar}Arabic results, ranked based on MAP, the official evaluation measure. The best score per evaluation measure is shown in bold.}
\end{table}

\textbf{Arabic.} Only two teams participated in the Arabic task~\cite{T1-Ghanem:2018,T1-Yasser:2018}, using basically the same models that they had for English. The \emph{bigIR}~\cite{T1-Yasser:2018} team translated automatically the test input to English and then ran their English system, while \emph{UPV--INAOE--Autoritas} translated to Arabic the English lexicons their representation was based on, and then trained an Arabic system on the Arabic training data, which they finally ran on the Arabic test input. It is worth noting that for English \emph{UPV--INAOE--Autoritas} outperformed \emph{bigIR}, but for Arabic it was the other way around. We suspect that a possible reason might be the direction of machine translation and also the presence/lack of context.
On one hand, translation into English tends to be better than into Arabic. Moreover, the translation of sentences is easier as there is context, whereas such a context is missing when translating lexicon entries in isolation. 

Finally, similarly to English, all runs managed to outperform the \emph{random} baseline by a margin, while the \emph{n-gram} baseline was strong yet possible to beat.

\section{Discussion}
\label{sec:discuss}

While the training data included debates only, the test data also contained speeches. Thus, it is interesting to see how systems perform on debates vs. speeches. Table~\ref{tab:results-checkworthines-deb-speech} shows the MAP for the primary submissions for both English and Arabic. Interestingly, speeches turn out to be easier than debates. We are not sure why this should be the case, but it might be because the speeches in our test dataset have about twice as many check-worthy claims as there are in the debates (see Table~\ref{tab:datasets-overview}).

We further experimented with constructing an ensemble using the scores by the individual systems. In particular, we first performed min-max normalization of the predictions of the individual systems, and then we summed these normalized scores.\footnote{We also tried summing the reciprocal ranks of the rankings that the systems assigned to each sentence, but this yielded much worse results.} 
The results are shown in Table~\ref{tab:results-checkworthines-ablation}. We can see that there is small improvement for the ensemble over the best individual system in terms of MAP for both English and Arabic. The results for the other evaluation measures are somewhat mixed for English, but there is clear improvement for Arabic. 

Table~\ref{tab:results-checkworthines-ablation} further shows the results for ablation experiments, where we remove one system from the ensemble. We can see that in most cases, removing an individual system yields lower MAP. A notable exception is \emph{blue}, removing which yields improvements in terms of MAP and some other evaluation measures. Moreover, we can see that different ablations can improve over any of the evaluation measures. This suggests that there is potential for improving the overall results by combining the approaches used by the different teams; this should be also possible at the feature/model level.

\begin{table}[t]
\centering 
\begin{tabular}{r l@{\hspace{1mm}} c@{\hspace{1mm}}c}
\toprule
\multicolumn{4}{c}{\bf English}\\
\multicolumn{2}{c}{\bf Team}	& \bf Debates	& \bf Speeches	\\	
\midrule
\cite{T1-Zuo:2018}	& Prise de Fer	& .1011$_{(1)}$	& .1460$_{(1)}$	\\
\cite{T1-Hansen:2018}	& Copenhagen	& .0757$_{(2)}$	& .1310$_{(3)}$	\\
\cite{T1-Ghanem:2018}	& UPV--INAOE--Aut.& .0521$_{(4)}$	& .1373$_{(2)}$	\\
\cite{T1-Yasser:2018}	& bigIR		& .0693$_{(3)}$	& .1290$_{(4)}$	\\
			& fragarach	& .0512$_{(5)}$	& .0932$_{(5)}$	\\
			& blue		& .0506$_{(6)}$	& .0920$_{(6)}$	\\
\cite{T1-Agez:2018}	& RNCC		& .0417$_{(7)}$	& .0717$_{(7)}$	\\	
\bottomrule
\\
\end{tabular}
\begin{tabular}{r l@{\hspace{1mm}} c@{\hspace{1mm}}c}
\toprule
\multicolumn{4}{c}{\bf Arabic}\\
\multicolumn{2}{c}{\bf Team}	& \bf Debates	& \bf Speeches	\\	
\midrule
\cite{T1-Yasser:2018}	& bigIR		& .0650$_{(1)}$	& .1397$_{(1)}$	\\
\cite{T1-Ghanem:2018}	& UPV--INAOE--Aut.& .0461$_{(2)}$	& .0834$_{(2)}$	\\	
\bottomrule
\end{tabular}
\caption{\label{tab:results-checkworthines-deb-speech}MAP for the primary submissions for debates vs. speeches.}
\end{table}

\begin{table}[t]
\centering 
\begin{tabular}{l @{\hspace{1mm}} c@{\hspace{1mm}}c@{\hspace{1mm}} c@{\hspace{1mm}}  c@{\hspace{1mm}} c@{\hspace{1mm}}c c@{\hspace{1mm}} c@{\hspace{1mm}} c}
\toprule
 		 & &\multicolumn{5}{c}{\bf ENGLISH}	\\
		& \bf  MAP	& MRR		& MR-P		& MP@1		& MP@3		& MP@5		& MP@10		& MP@20		& MP@50  \\
\midrule
\bf Best team: \emph{Prise de Fer} & .1332	& \bf .4965	&  .1352	& \bf \underline{.4286}	& \bf \underline{.2857}	& .2000	& .1429	&  .1571 & .1200\\ 

\midrule

\bf Ensemble: \emph{SUM scores} & \bf .1378 & 	.4479 & 	\bf .1726 & 	.2857 & 	.2381 & 	.2000 & 	\bf .2000 & 	.1571 & 	.1200\\

\,\,\,$-$\textbf{blue} & \underline{.1437} & 	.4533 & 	\underline{.1839} & 	.2857 & 	.2381 & 	\underline{.2571} & 	.2000 & 	\underline{.2000} & 	\underline{.1286}\\
\,\,\,$-$\textbf{Prise de Fer} & .1341 & 	.3890 & 	.1537 & 	.1429 & 	.2381 & 	.2286 & 	.2000 & 	.1571 & 	.1171\\
\,\,\,$-$\textbf{Copenhagen} & .1322 & 	.4449 & 	.1473 & 	.2857 & 	.2381 & 	.2286 & 	.2143 & 	.1357 & 	.1200\\
\,\,\,$-$\textbf{fragarach} & .1302 & 	.3888 & 	.1574 & 	.1429 & 	.2381 & 	.2000 & 	\underline{.2286} & 	.1500 & 	.1257\\
\,\,\,$-$\textbf{RNCC} & .1298 & 	.3885 & 	.1596 & 	.1429 & 	.2381 & 	.2286 & 	.2143 & 	.1500 & 	.1171\\
\,\,\,$-$\textbf{UPV-INAOE-Autoritas} & .1257 & 	.4545 & 	.1466 & 	.2857 & 	\underline{.2857} & 	.1714 & 	.1857 & 	.1429 & 	.1200\\
\,\,\,$-$\textbf{bigIR} & .1205 & 	\underline{.5250} & 	.1195 & 	\underline{.4286} & 	.2381 & 	.2286 & 	.1857 & 	.1357 & 	.1114\\
\bottomrule
\\
\toprule
 		 & &\multicolumn{5}{c}{\bf ARABIC}	\\
		& \bf  MAP	& MRR		& MR-P		& MP@1		& MP@3		& MP@5		& MP@10		& MP@20		& MP@50  \\
\midrule
\bf Best team: \emph{bigIR} & .0899	& .1180	& .1105	& .0000	& .0000	& .0000	& .1333	& .1000	& .1133\\ 

\midrule

\bf Ensemble: \emph{SUM scores} & \bf .0931 & 	\bf .4083 & 	.1105 & 	\bf .3333 & 	\bf .1111 & 	\bf .0667 & 	.1333 & 	\bf .1167 & 	\bf .1200\\

\bottomrule
\end{tabular}
\caption{\label{tab:results-checkworthines-ablation}Ablation results for an ensemble summing the  participating systems, as well as for ablation excluding each of the systems from the ensemble.}
\end{table}

\section{Conclusion and Future Work}
\label{sec:conclusions}

We provided an overview of the 
CLEF-2018 CheckThat! Lab on Automatic Identification and Verification of Political Claims, with focus on Task 1: Check-Worthiness, which asked to predict which claims in a political debate should be prioritized for fact-checking. We offered the task in both English and Arabic.

Our evaluation framework consisted of a dataset of five debates and five speeches divided into training and testing set, and a MAP-based evaluation. A total of thirty teams registered to participate in the Lab and seven teams actually submitted systems for Task~1. The most successful approaches used by the participants relied on recurrent and multi-layer neural networks, as well as on combinations of distributional representations, on matchings claims' vocabulary against lexicons, and on measures of syntactic dependency. The best systems achieved mean average precision of 0.18 and 0.15 on the English and on the Arabic test datasets, respectively. This leaves large room for further improvement, and thus we release\footnote{\url{http://alt.qcri.org/clef2018-factcheck/}} all datasets and the scoring scripts, which should enable further research in check-worthiness estimation.

In future iterations of the lab, we plan to add more debates and speeches, both annotated and unannotated, which would enable semi-supervised learning. We further want to add annotations for the same debates/speeches from different fact-checking organizations, which would allow using multi-task learning \cite{gencheva-EtAl:2017:RANLP}.

\section*{Acknowledgments}
This work was made possible in part by NPRP grant\# NPRP 7-1313-1-245 from the Qatar National Research Fund (a member of Qatar Foundation). Statements made herein are solely the responsibility of the authors. 

\bibliographystyle{splncs04}
\bibliography{clef18_checkthat.bib,otherrefs.bib}

\end{document}